\documentclass[runningheads]{llncs}
\usepackage{graphicx}
\usepackage{color}
\usepackage{amsmath}
\usepackage{amssymb}

\usepackage{amsthm}

\newtheorem{thm}{Theorem}
\newtheorem{rem}{Remark}
\theoremstyle{remark}
\newtheorem*{proo}{Proof}
\newtheorem*{counterexample}{Counterexample}

\usepackage{bm}
\usepackage{algcompatible}

\usepackage{algorithm,algpseudocode}
\usepackage{multirow}
\usepackage{booktabs}
\usepackage{xcolor}

\begin{document}
\title{Wasserstein Divergence for GANs} 

\titlerunning{Wasserstein Divergence for GANs}
%
\author{Jiqing Wu\inst{1} \and
Zhiwu Huang\inst{1} \and
Janine Thoma\inst{1} \and
Dinesh Acharya\inst{1} \and \\ 
Luc Van Gool\inst{1,2}}
%
\authorrunning{Wu et al.}
%
\institute{Computer Vision Lab, ETH Zurich, Switzerland
\email{\{jwu,zhiwu.huang,jthoma,vangool\}@vision.ee.ethz.ch, acharyad@student.ethz.ch}\\
\and
VISICS, KU Leuven, Belgium}
%
\maketitle              
\begin{abstract}
In many domains of computer vision, generative adversarial networks (GANs) have achieved great success, among which the family of Wasserstein GANs (WGANs) is considered to be state-of-the-art due to the theoretical contributions and competitive qualitative performance.
However, it is very challenging to approximate the $k$-Lipschitz constraint required by the Wasserstein-1 metric~(W-met). 
In this paper, we propose a novel Wasserstein divergence~(W-div), which is a relaxed version of W-met and does not require the $k$-Lipschitz constraint.
As a concrete application, we introduce a Wasserstein divergence objective for GANs~(WGAN-div),
which can faithfully approximate W-div through optimization.
Under various settings, including progressive growing training, we demonstrate the stability of the proposed WGAN-div owing to its theoretical and practical advantages over WGANs.
Also, we study the quantitative and visual performance of WGAN-div on standard image synthesis benchmarks,
showing the superior performance of WGAN-div compared to the state-of-the-art methods.

\end{abstract}

\section{Introduction}
Over the past few years, we have witnessed the great success of generative adversarial networks (GANs)~\cite{goodfellow2014generative} for a variety of applications. GANs are a useful family of generative models that expresses generative modeling as a zero-sum game between two networks: A generator network produces plausible samples given some noise, while a discriminator network distinguishes between the generator's output and real data.
There are numerous works inspired by the original GANs,~\cite{radford2015unsupervised,berthelot2017began,mao2016least,zhao2016energy} to name a few.
While GANs can produce visually pleasing samples, they lack a reliable way of measuring the difference between fake and real data distribution, which leads to unstable training.

To address this issue,
\cite{arjovsky2017wasserstein} introduced the Wassestein-1 metric (W-met) to the GAN framework.
Compared to the Jensen-Shannon (JS) or the Kullback-Leibler (KL) divergence,
W-met is considered to be more sensible for distributions supported by low dimensional manifolds. 
Given that the primal form of W-met is intractable to compute,
\cite{arjovsky2017wasserstein} proposed to use the dual form of W-met, which requires the $k$-Lipschitz constraint.
A series of ideas
\cite{arjovsky2017wasserstein,gulrajani2017improved,miyato2018spectral,wei2018improve} were proposed to approximate the dual W-met and achieved impressive results compared to the non-Wasserstein based GANs.
However, they generally suffer from unsatisfying regularization for the $k$-Lipschitz constraint,
mainly because it is a very strict constraint and non-trivial to approximate~\cite{wei2018improve,roth2017stabilizing}. 

Other studies have tackled the stability issue from different angles.
For example,
\cite{roth2017stabilizing} proposed a gradient-based regularizer associated with the $\mathfrak{f}$-divergence~\cite{nowozin2016f} to address the dimensional misspecification.
In order to stabilize the training towards high resolution images,
\cite{zhang2016stackgan,huang2016stacked} applied deep stack architectures by incorporating extra information. 
Recently, building upon the dual W-met objective of~\cite{gulrajani2017improved},
\cite{karras2017progressive} presented a sophisticated progressive growing training scheme and obtained excellent high resolution images.

In this paper, we propose to resolve the $k$-Lipschitz constraint by introducing a relaxed version of W-met and incorporating it in the GAN framework. Our contributions can be summarized as follows:
\begin{enumerate}
\item We introduce a novel Wasserstein divergence (W-div) and prove that the proposed W-div is a symmetric divergence. Moreover, we explore the connection between the proposed W-div and W-met.
\item Benefiting from the non-challenging constraint required by the W-div,  we introduce Wasserstein divergence GANs (WGAN-div) as its practical application.
The proposed objective can faithfully approximate the corresponding W-div through optimization.
\item We demonstrate the stability of WGAN-div under various settings including progressive growing training. Also, we conduct various experiments on standard image synthesis benchmarks and present superior results of WGAN-div compared to the state-of-the-art methods, both quantitatively and qualitatively.
\end{enumerate}

\section{Background}


Imagine there are two players in a game.
One player (Generator) intends to generate visually plausible images, aiming to fool its opponent,
while the opponent (Discriminator) attempts to discriminate real images from synthetic images.
Such adversarial competition is the key idea behind GAN models. 
To measure the distance between real and fake data distributions, \cite{goodfellow2014generative} proposed the objective 
\begin{equation}
\label{eq:gan}
L_{\text{JS}}(\mathbb{P}_r, \mathbb{P}_g) = \underset{\bm{x} \sim \mathbb{P}_r}{\mathbb{E}}[\text{ln}(f(\bm{x}))] + \underset{\tilde{\bm{x}} \sim \mathbb{P}_g}{\mathbb{E}}[\text{ln}(1 -f(\tilde{\bm{x}}))],
\end{equation}
which can be interpreted as the JS divergence up to a constant~\cite{arjovsky2017towards} and where $f$ is a discriminative function.
The model can thus be defined as a min-max optimization problem:
\begin{equation}
\label{eq:minimax}
\underset{G}{\mathrm{min}} \underset{D}{\mathrm{max}} \underset{\bm{x} \sim \mathbb{P}_r}{\mathbb{E}}[\mathrm{ln}(D(\bm{x}))] + \underset{G(\bm{z}) \sim \mathbb{P}_g}{\mathbb{E}}[\mathrm{ln}(1 - D(G(\bm{z})))],
\end{equation}
where $G$ is the generator parametrized by a neural network and $D$ is the discriminative neural network parametrizing $f$.
Usually, we let $\bm{z}$ be low dimensional random noise, 
and $\bm{x}, G(\bm{z})$ are the real and fake data satisfying the probability measures $\mathbb{P}_r, \mathbb{P}_g$.

\smallskip
\noindent \textbf{Wasserstein GANs (WGANs).}
The rise of the Wasserstein-1 metric (W-met) in GAN models is primarily motivated by unstable training caused by the gradient vanishing problem~\cite{arjovsky2017wasserstein}.
Given two probability measures $\mathbb{P}_r, \mathbb{P}_g$, the W-met~\cite{villani2008optimal} is defined as
\begin{equation}
\label{eq:wass}
\mathcal{W}_1(\mathbb{P}_r, \mathbb{P}_g) = \underset{f \in \mathrm{Lip}_1}{\mathrm{sup}} \underset{\bm{x} \sim \mathbb{P}_r}{\mathbb{E}}[f(\bm{x})] - \underset{\tilde{\bm{x}} \sim \mathbb{P}_{g}}{\mathbb{E}}[f(\tilde{\bm{x}})],
\end{equation}
where $\mathrm{Lip}_1$ is the function space of all $f$ satisfying the 1-Lipschitz constraint $\|f\|_L \leq 1$. It is worth mentioning that $\mathcal{W}_1$ is invariant up to a positive scalar $k$ if the Lipschitz constraint is modified to be $k$. $\mathcal{W}_1$ is believed to be more sensible to distributions supported by low dimensional manifolds such as image, video, etc.
Generally, the existing Wasserstein GANs (WGANs) fall into two categories:

\noindent \emph{Weight Constraints.}
To approximately satisfy the Lipschitz constraint, \cite{arjovsky2017wasserstein} proposed a weight clipping method that imposes a hard threshold $c > 0$ on the weights $\bm{w}$ of the discriminator $D$, which parametrizes $f$ in Eq.~\ref{eq:wass}:
\begin{equation}
\label{eq:clip}
\bm{w'} =
  \begin{cases}
    \bm{w}       & \quad  \text{if } |\bm{w}| < c\\
    c  & \quad \text{if } \bm{w} \geq c \\
    -c & \quad \text{if } \bm{w} \leq -c 
  \end{cases}
\end{equation}
This approach was proven to be unsatisfactory by~\cite{gulrajani2017improved}, 
since through weight clipping, the neural network tends to learn oversimplified functions. 
Later, \cite{miyato2018spectral} proposed spectral normalization GANs (SNGANs). 
To impose the $1$-Lipschitz constraint, SNGANs normalize the weights $\bm{w}_i$ of each layer $i$ by the $L_2$ matrix norm, 
\begin{equation}
\label{eq:spectral}
\bm{w'}_i = \frac{\bm{w}_i}{\|\bm{w}_i \|_2}.
\end{equation}
Because the set of functions satisfying the local $1$-Lipschitz constraint is merely a subset of the function space $\text{Lip}_1$, such a constraint inevitably narrows the effective search space and entails a sub-optimal solution.

\noindent \emph{Gradient Constraints.} 
To overcome the disadvantages of weight clipping, \cite{gulrajani2017improved} introduced a gradient penalty term to Wasserstein GANs (WGAN-GP). The objective is defined as 
\begin{equation}
\label{eq:wass1}
L_{\text{GP}} = \underbrace{\underset{\bm{x} \sim \mathbb{P}_r}{\mathbb{E}}[f(\bm{x})] - \underset{\tilde{\bm{x}} \sim \mathbb{P}_{g}}{\mathbb{E}}[f(\tilde{\bm{x}})]}_\text{Wasserstein term} 
+ k \underbrace{\underset{\Hat{\bm{x}} \sim \mathbb{P}_{y}}{\mathbb{E}}[(\|\nabla f(\Hat{\bm{x}})\|_2 - 1) ^2]}_\text{gradient penalty},
\end{equation}
where $\nabla$ is the gradient operator and $\mathbb{P}_{y}$ is the distribution obtained by sampling uniformly along straight lines between points from the real and fake data distributions $\mathbb{P}_{r}$ and $\mathbb{P}_{g}$. 
As pointed out by~\cite{wei2018improve,roth2017stabilizing}, with a finite number of training iterations on limited input samples, it is very difficult to guarantee the $k$-Lipschitz constraint for the whole input domain.
Thus, \cite{wei2018improve} further proposed Wasserstein GANs with a consistency term~(CTGANs). Inspired by the original $1$-Lipschitz constraint, CTGANs add the following term to Eq.~\ref{eq:wass1},
\begin{equation}
\label{eq:ct}
\text{CT}|_{\bm{x}_1, \bm{x}_2} = \mathbb{E}_{\bm{x}_1, \bm{x}_2}[\textrm{max}(0, \frac{d(f(\bm{x}_1), f(\bm{x}_2))}{d(\bm{x}_1, \bm{x}_2)} - c)],
\end{equation}
where $\bm{x}_1, \bm{x}_2$ are two data points, $d$ is a metric and $c$ is a threshold.
Recently, to improve stability and image quality,
\cite{karras2017progressive} proposed a training scheme in which GANs are grown progressively.
In addition to progressive growing, \cite{karras2017progressive} also proposed an objective $L_\text{PG} = L_\text{GP} + \text{PG}$, where
\begin{equation}
\label{eq:pg}
\text{PG} =
  \begin{cases}
    \underset{\Hat{\bm{x}} \sim \mathbb{P}_{y}}{\mathbb{E}}[(\|\nabla f(\Hat{\bm{x}})\|_2 - 750) ^2 / 750 ^2]  &  \quad  \text{for CIFAR-10} \\
    0.001 \underset{\Hat{\bm{x}} \sim \mathbb{P}_{y}}{\mathbb{E}}[\|\nabla f(\Hat{\bm{x}})\|_2 ^2]       & \quad  \text{for other datasets}
  \end{cases}
\end{equation}
\noindent \textbf{$\mathfrak{f}$-GANs.}
Outside the family of Wasserstein metrics, there is another important family of divergences---the $\mathfrak{f}$-divergences.
\cite{nowozin2016f} argued that $\mathfrak{f}$-divergence can be used for training
generative samplers and proposed $\mathfrak{f}$-GANs. Since the $\mathfrak{f}$-GANs are vulnerable to the dimension mismatch between fake and real data, \cite{roth2017stabilizing} proposed a gradient-based regularizer to stabilize the training and gave an example based on JS-divergence:
\begin{equation}
\begin{aligned}
L_{\text{RJS}}(\mathbb{P}_r, \mathbb{P}_g) &= \underset{\bm{x} \sim \mathbb{P}_r}{\mathbb{E}}[\text{ln}(f(\bm{x}))] + \underset{\tilde{\bm{x}} \sim \mathbb{P}_g}{\mathbb{E}}[\text{ln}(1 -f(\tilde{\bm{x}}))] - k \Omega(\mathbb{P}_r, \mathbb{P}_g)  \\
\Omega(\mathbb{P}_r, \mathbb{P}_g)  &:= \underset{\bm{x} \sim \mathbb{P}_r}{\mathbb{E}} \left[ (1-f(\bm{x}))^2 \| \nabla f(\bm{x}) \|^2 \right] +  \underset{\tilde{\bm{x}} \sim \mathbb{P}_g}{\mathbb{E}} \left[f(\tilde{\bm{x}})^2 \| \nabla f(\tilde{\bm{x}}) \|^2 \right]. 
\label{eq:regularizer}
\end{aligned}
\end{equation}

\noindent \textbf{Information Geometry.}
In information geometry, \cite{karakida2017information} studied the connections between the Wasserstein distance and the Kullback-Leibler (KL) divergence employed by early GANs. They exploit the fact that by regularizing the Wasserstein distance with entropy, the entropy relaxed Wasserstein distance introduces a divergence and naturally defines certain geometrical structures from the information geometry viewpoint.

\section{Proposed Method}
As discussed above, it is very challenging to approximate the W-met. This is due to the gap between limited input samples on the one hand and the strict $1$-Lipschitz constraint on the whole input sample domain~\cite{Rothe-IJCV-2016,wei2018improve} on the other hand. 
At the same time, it is natural to ask whether there exists an optimal $f^*$ for W-met~(Eq.~\ref{eq:wass}). 
According to~\cite{evans1997partial}, by solving a family of minimization problems given $p > 0$ 
\begin{equation}
f_p = \underset{f \in W_c^{1, p}}{\mathrm{argmin}}  \underset{\bm{x} \sim \mathbb{P}_r}{\mathbb{E}}[f(\bm{x})] - \underset{\tilde{\bm{x}} \sim \mathbb{P}_{g}}{\mathbb{E}}[f(\tilde{\bm{x}})]
+ \frac{1}{p} \underset{\Hat{\bm{x}} \sim \mathbb{P}_{u}}{\mathbb{E}}[\|\nabla f(\Hat{\bm{x}})\|^p],
\label{eq:exist}
\end{equation}
where $\mathbb{P}_u$ is a Radon probability measure and $W_c^{1, p}$ is the Sobolev space containing all the functions $f$ in $L^p$ space with first order weak derivatives and compact support, we can find a sequence $p_k \rightarrow \infty$ such that $f_{p_k} \rightarrow -f^*$.

\subsection{Wasserstein Divergence}
The connection between Eq.~\ref{eq:exist} and W-met inspires us to propose a novel Wasserstein divergence~(W-div) and we prove that it is indeed a valid symmetric divergence.
\begin{thm}(Wasserstein divergence)
\label{thm:theorem1}
Let $\Omega \subset \mathbb{R}^n$ be an open, bounded, connected set and S be the set of all the Radon probability measures on $\Omega$. 
If for some $p \neq 1, k > 0$ we define 
\begin{equation}
\label{eq:wd}
\begin{aligned}
\mathcal{W}_{p,k}^{'}:S \times S &\rightarrow \mathbb{R}^- \cup \{ 0 \} \\
(\mathbb{P}_r, \mathbb{P}_g) &\rightarrow 
\underset{f \in C_c^{1}(\Omega)}{\mathrm{inf}}  \underset{\bm{x} \sim \mathbb{P}_r}{\mathbb{E}}[f(\bm{x})] - \underset{\tilde{\bm{x}} \sim \mathbb{P}_{g}}{\mathbb{E}}[f(\tilde{\bm{x}})]
+ k \underset{\Hat{\bm{x}} \sim \mathbb{P}_{u}}{\mathbb{E}}[\|\nabla f(\Hat{\bm{x}})\|^p],
\end{aligned}
\end{equation}
where $C_c^1(\Omega)$ is the function space of all the first order differentiable functions on $\Omega$ with compact support,
then $\mathcal{W}_{p,k}^{'}$ is a symmetric divergence (up to the negative sign).
\end{thm}
\begin{proo}
See supplementary material.
\end{proo}
By imposing the $C_c^1(\Omega)$ function space, we rule out pathological functions with weak derivatives.
Compared to the $k$-Lipschitz constraint, $f \in C_c^1(\Omega)$ is less restrictive, since $\|\nabla f\|$ does not need to be bounded by a hard threshold $k$. 
Given the universal approximation theorem and the modern architecture of neural networks---stacking differentiable layers to form a nonlinear differentiable function---$f \in C_c^1(\Omega)$ can easily be parameterized by a neural network.

In the following we further explore the connection between the proposed W-div and the original W-met in Eq.~\ref{eq:wass}. 
\begin{rem}(Upper bound)
\label{rem2}
Given Radon probability measures $\mathbb{P}_r, \mathbb{P}_g, \mathbb{P}_u$ on $\Omega$, let
\begin{equation}
\label{eq:wd2}
\mathcal{W}_{\mathbb{P}_u}^{'}(\mathbb{P}_r, \mathbb{P}_g):= 
\underset{f \in C_c^{\infty}(\Omega)}{\mathrm{inf}}  \underset{\bm{x} \sim \mathbb{P}_r}{\mathbb{E}}[f(\bm{x})] - \underset{\tilde{\bm{x}} \sim \mathbb{P}_{g}}{\mathbb{E}}[f(\tilde{\bm{x}})]
+ \frac{1}{2} \underset{\Hat{\bm{x}} \sim \mathbb{P}_{u}}{\mathbb{E}}[(\|\nabla f(\Hat{\bm{x}})\|^2],
\end{equation}
where $C_c^{\infty}$ is the function space of all the smooth functions $f$ with compact support.
There exists an optimal $f^*$ for $\mathcal{W}_1$(Eq.~\ref{eq:wass}) such that 
\begin{equation}
\label{eq:solv}
\mathcal{W}_1(\mathbb{P}_r, \mathbb{P}_g) =  \underset{\bm{x} \sim \mathbb{P}_r}{\mathbb{E}}[f^*(\bm{x})] - \underset{\tilde{\bm{x}} \sim \mathbb{P}_{g}}{\mathbb{E}}[f^*(\tilde{\bm{x}})],
\end{equation}
and a $\mathcal{W}_{\mathbb{P}_{u^*}}^{'}$ determined by $f^*$ such that
\begin{equation}
\label{eq:wd2_1}
\mathcal{W}_{\mathbb{P}_{u^*}}^{'}(\mathbb{P}_r, \mathbb{P}_g) = \underset{\mathbb{P}_{u} \in S}{\mathrm{sup}} \,  \mathcal{W}_{\mathbb{P}_{u}}^{'}(\mathbb{P}_r, \mathbb{P}_g). 
\end{equation}
Please see the detailed discussion in~\cite{evans1997partial}.
\end{rem}
Remark~\ref{rem2} indicates that $\mathcal{W}_{\mathbb{P}_{u^*}}^{'}$, which is determined by the optimal $f^*$, is the upper bound of our W-div $\mathcal{W}_{\mathbb{P}_{u}}^{'}$\footnote{$\mathcal{W}_{\mathbb{P}_{u}}^{'}$ is a family of special cases of Eq.~\ref{eq:wd} with a more restrictive function space $C_c^{\infty}$.}. 

Given the similarities between our proposed W-div and $L_{\text{GP}}$ (Eq.~\ref{eq:wass1}), it may be interesting to know if there exists a divergence corresponding to $L_{\text{GP}}$. In general, the answer is no.
\begin{rem}
\label{rem}
If for $n > 0$ we let 
\begin{equation}
\label{eq:wd1}
\mathcal{W}_{p,k,n}^{''}(\mathbb{P}_r, \mathbb{P}_g):= 
\underset{f \in C_c^{1}(\Omega)}{\mathrm{inf}}  \underset{\bm{x} \sim \mathbb{P}_r}{\mathbb{E}}[f(\bm{x})] - \underset{\tilde{\bm{x}} \sim \mathbb{P}_{g}}{\mathbb{E}}[f(\tilde{\bm{x}})]
+ k \underset{\Hat{\bm{x}} \sim \mathbb{P}_{u}}{\mathbb{E}}[(\|\nabla f(\Hat{\bm{x}})\| - n)^p],
\end{equation}
then $\mathcal{W}_{p,k,n}^{''}$ is \textbf{not} a divergence in general.
\end{rem}
\begin{counterexample} 
Assuming $\Omega = (-1, 1)$ and $p = 2$, it suffices to show that $\mathcal{W}_{2,k,n}^{''}(\mathbb{P}_r, \mathbb{P}_g) \neq 0$ for $\mathbb{P}_r = \mathbb{P}_g$ almost everywhere. Since ${\mathbb{E}}_{\bm{x} \sim \mathbb{P}_r}[f(\bm{x})]$ and ${\mathbb{E}}_{\tilde{\bm{x}} \sim \mathbb{P}_{g}}[f(\tilde{\bm{x}})]$ cancel out, in order to guarantee $\mathcal{W}_{2,k,n}^{''}(\mathbb{P}_r, \mathbb{P}_g) = 0$, $\|\nabla f(\hat{\bm{x}})\|$ must be equal to $n$ on $(-1, 1)$, 
which implies that $f$ is affine and contradicts the compact support constraint.
For $m$-dimensional sets such as $(-1, 1)^m$ and an even integer $p$ we need to employ the uniqueness argument of the Picard-Lindel{\"o}f Theorem to show that $f$ can only be affine.
\end{counterexample}
Remark~\ref{rem} implies that the plausible statistic distance $\mathcal{W}_{2,k,1}^{''}$  corresponding to Eq.~\ref{eq:wass1} is neither a divergence, nor a valid metric.

\subsection{Wasserstein Divergence GANs}
Although W-met enjoys the tempting property of providing useful gradients,
in practice, the original formulation ${\mathbb{E}}_{\bm{x} \sim \mathbb{P}_r}[f(\bm{x})] - {\mathbb{E}}_{\tilde{\bm{x}} \sim \mathbb{P}_{g}}[f(\tilde{\bm{x}})]$ of W-met cannot be directly applied as an objective without imposing the strict $1$-Lipschitz constraint. 
In contrast, it is very straightforward to use our proposed W-div as an objective.
Therefore, we introduce Wasserstein divergence GANs (WGAN-div). Our objective can be smoothly derived as
\begin{equation}
\label{eq:obj1}
L_{\text{DIV}} = \underset{\bm{x} \sim \mathbb{P}_r}{\mathbb{E}}[f(\bm{x})] - \underset{\tilde{\bm{x}} \sim \mathbb{P}_{g}}{\mathbb{E}}[f(\tilde{\bm{x}})]
+ k \underset{\Hat{\bm{x}} \sim \mathbb{P}_{u}}{\mathbb{E}}[\|\nabla f(\Hat{\bm{x}})\|^p],
\end{equation}
which is identical to the formulation of W-div without the infimum.
Minimizing $L_{\text{DIV}}$ faithfully approximates $\mathcal{W}_{p,k}^{'}$, in a sense that the decrease of $L_{\text{DIV}}$ indicates a better approximation of $\mathcal{W}_{p,k}^{'}$.
In comparison, lowering $L_{\text{GP}}$ does not necessarily imply that $L_{\text{GP}}$ approximates $\mathcal{W}_1$ better, since $L_{\text{GP}}$ can be decreased at the cost of violating the gradient penalty term (Eq.~\ref{eq:wass1}).

\begin{algorithm}
\caption{The proposed WGAN-div algorithm}
\begin{algorithmic}[1]
\Require{Batch size $m$, generator $G$ and discriminator $D$, power $p$, coefficient $k$,
    training iterations $n$, and other hyperparameters}
\For{$i \gets 1 \textrm{ to } n$} 
    \State Sample real data $\bm{x}_1, \ldots ,\bm{x}_m $ from $\mathbb{P}_r$
    \State Sample Gaussian noise $\bm{z}_1, \ldots ,\bm{z}_m $ from $\mathcal{N}(0,1)$
    \State Sample vector $\bm{\mu} = ( \mu_1, \ldots ,\mu_m )$ from uniform distribution $U[0, 1]$ such that 
    \State $\bm{\hat{x}}_j = (1- \mu_{j}) \bm{x}_j + \mu_{j} G(\bm{z}_j) $ 
    \State Update the weights $\bm{w}_G$ of $G$ by descending:  
    
           $\bm{w}_G \leftarrow \mathrm{Adam}(\nabla_{\bm{w}_G}( \frac{1}{m} \sum_{j = 1} ^m D(G(\bm{z}_j))), \bm{w}_G, \alpha, \beta_1, \beta_2)$
    \State Update the weights $\bm{w}_D$ of $D$ by descending:  
    
            $\bm{w}_D \leftarrow \mathrm{Adam}(\nabla_{\bm{w}_D}(\frac{1}{m} \sum_{j = 1} ^m  D(\bm{x}_j) - D(G(\bm{z}_j)) $ 
            
            $+ k \|\nabla_{\bm{\hat{x}}_j} D(\bm{\hat{x}}_j)\|^p), \bm{w}_D, \alpha, \beta_1, \beta_2)$
          
\EndFor
\end{algorithmic}
\label{alg}
\end{algorithm}

By incorporating our objective $L_{\text{DIV}}$ in the GAN framework, together with parameterizing $f \in C_c^1$ by a discriminator $D$ and the fake data distribution $\mathbb{P}_g$ by a generator $G$, 
our min-max optimization problem can be written as
\vspace{-0.2cm}
	\begin{equation}
	\min_{G} \max_{D}  \, \underset{G(\bm{z}) \sim \mathbb{P}_{g}}{\mathbb{E}}[D(G(\bm{z}))] - \underset{\bm{x} \sim \mathbb{P}_r}{\mathbb{E}}[D(\bm{x})] 
- k \underset{\Hat{\bm{x}} \sim \mathbb{P}_{u}}{\mathbb{E}}[\|\nabla_{\Hat{\bm{x}}} D(\Hat{\bm{x}})\|^p],
	\label{Eq3}
	\end{equation}
where $\bm{z}$ is random noise, $\bm{x}$ is the real data, and $\Hat{\bm{x}}$ is sampled as a linear combination of real and fake data points. For more studies of sampling strategies we refer readers to our supplementary material.
The final algorithm is obtained as shown in Alg.~\ref{alg}.
Following the good practice of~\cite{gulrajani2017improved}, 
our building blocks for $D$ and $G$ are residual blocks~\cite{he2016deep}.
The default architecture of WGAN-div is presented in Tab.~\ref{tab:net}.
We apply Adam optimization~\cite{kingma2014adam} to update $G$ and $D$. 
We study the crucial hyperparameters such as the coefficient $k$ and the power $p$ in the next section.

\begin{table}[!tp]
\centering
\begin{tabular}{cccc}
\toprule
\textbf{Generator}            &  Kernel size      & Resampling     & Output shape           \\ \midrule
Noise         & --            & --            & 128            \\
Linear          & --            & --            & $512\times4\times4$            \\
Residual block  & $[3\times3]\times2$    & Up           & $512\times8\times8$           \\
Residual block  & $[3\times3]\times2$    & Up           & $256\times16\times16$           \\
Residual block  & $[3\times3]\times2$    & Up           & $128\times32\times32$           \\
Residual block  & $[3\times3]\times2$    & Up           & $64\times32\times32$           \\
Conv, tanh  & $3\times3$    & --           & $3\times64\times64$           \\
\toprule

\textbf{Discriminator}            &          &         &            \\ \midrule
Conv            & $3\times3$    & --           & $64\times64\times64$ \\      
Residual block  & $[3\times3]\times2$    & Down           & $128\times32\times32$           \\
Residual block  & $[3\times3]\times2$    & Down           & $256\times16\times16$           \\
Residual block  & $[3\times3]\times2$    & Down           & $512\times8\times8$           \\
Residual block  & $[3\times3]\times2$    & Down         & $512\times4\times4$           \\

Linear   & --    & --           & 1    \\\toprule
\end{tabular}
\vspace{2mm}
\caption{The default architecture of WGAN-div for $64\times64$ image generation}
\label{tab:net}
\end{table}

\begin{table}[!tp]
\vspace{-0.7cm}
\centering
\resizebox{1\textwidth}{!}{
\begin{tabular}{ccc|ccc|ccc}
\toprule
&WGAN-GP & & &CTGAN & & &\textbf{WGAN-div} & \\ \midrule
\includegraphics[width=0.15\linewidth]{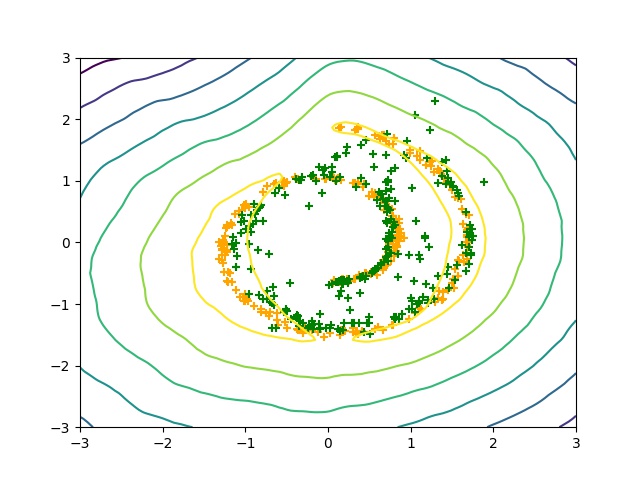}&
\includegraphics[width=0.15\linewidth]{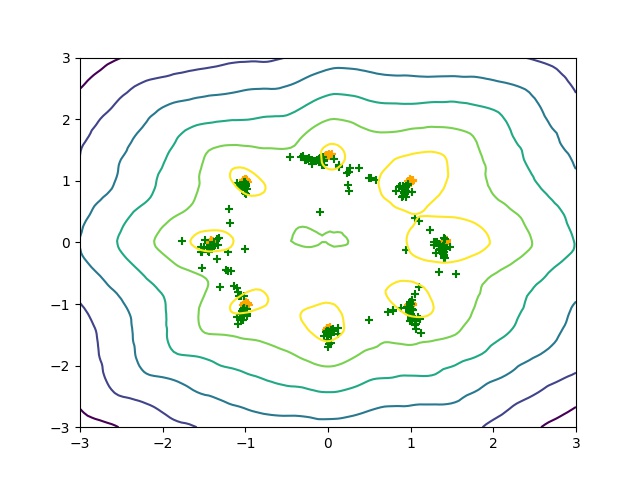}&
\includegraphics[width=0.15\linewidth]{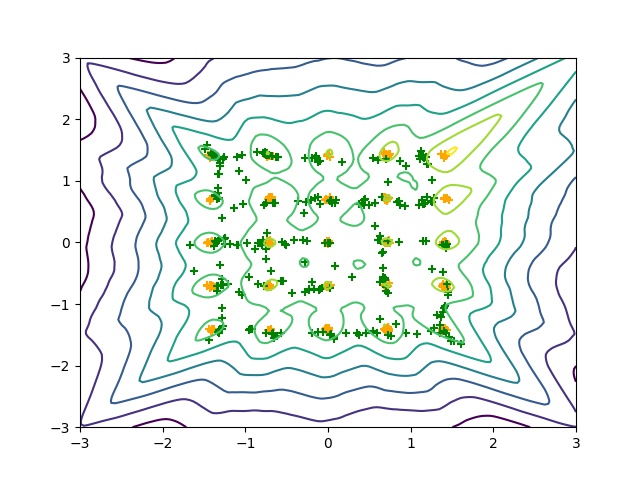}&

\includegraphics[width=0.15\linewidth]{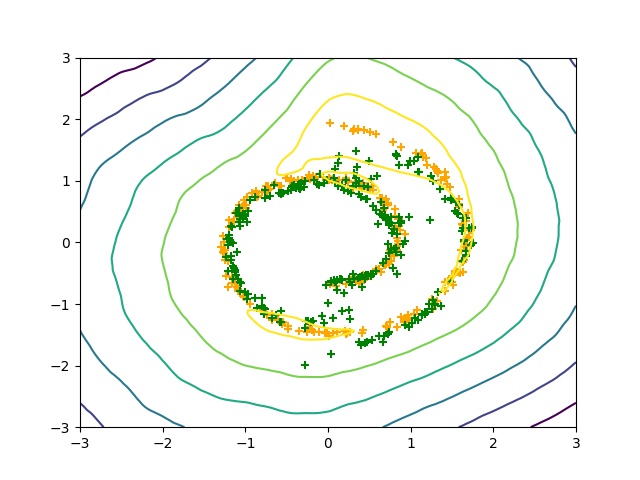}&
\includegraphics[width=0.15\linewidth]{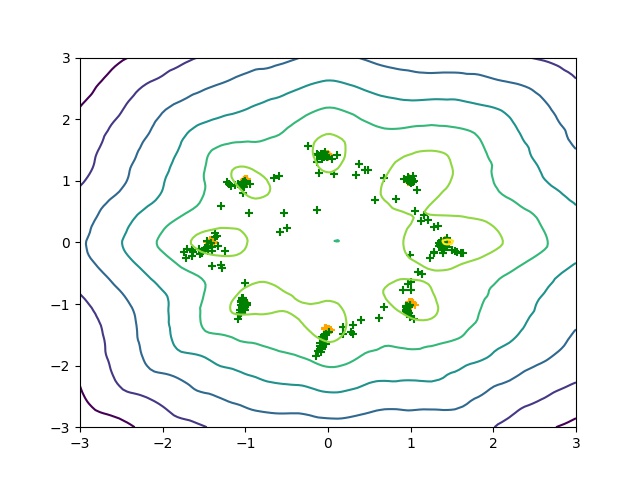}&
\includegraphics[width=0.15\linewidth]{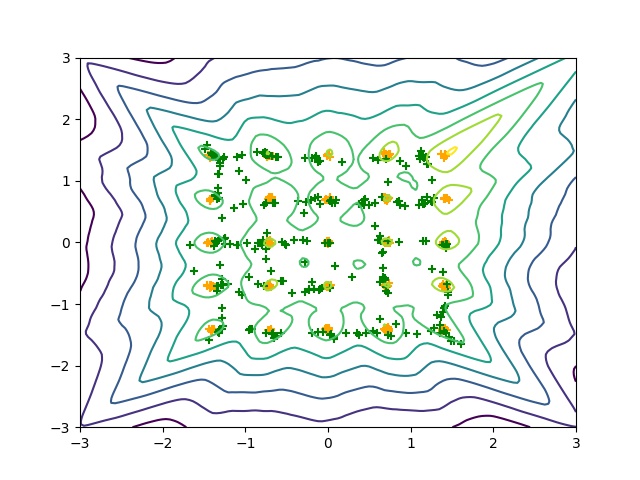}&

\includegraphics[width=0.15\linewidth]{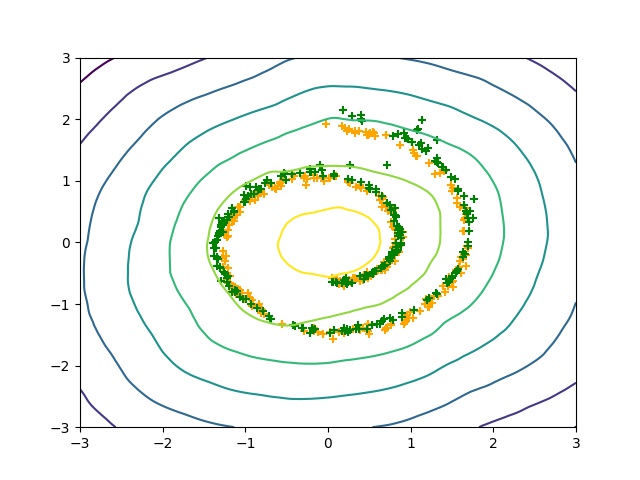}&
\includegraphics[width=0.15\linewidth]{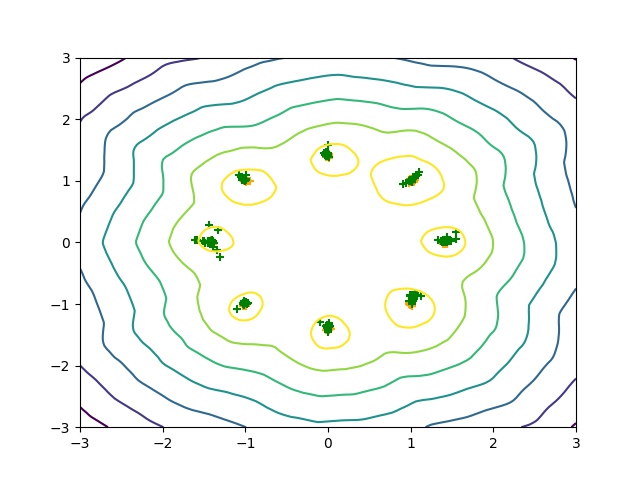}&
\includegraphics[width=0.15\linewidth]{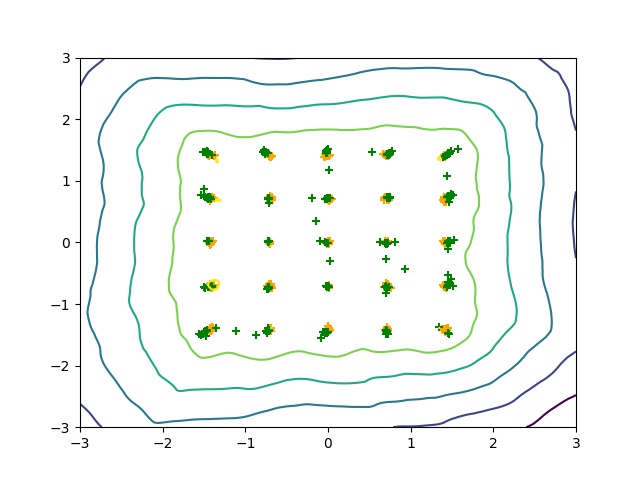}\\

0.02& 0.04 & 0.04 & 0.02 & 0.03 & 0.03  & \textbf{0.01} & \textbf{0.02} & \textbf{0.01}\\ \bottomrule
\end{tabular}
}
\vspace{2mm}
\caption{Visual and FID comparison for generated samples (green dots) and real samples (yellow dots) on Swiss Roll, 8 Gaussians and 25 Gaussians. The value surfaces of the discriminators are also plotted.}
\label{tab:3toy}
\end{table}

\begin{figure}[tb]
\centering
\begin{tabular}{c}
\includegraphics[width=0.9\linewidth, height=5cm]{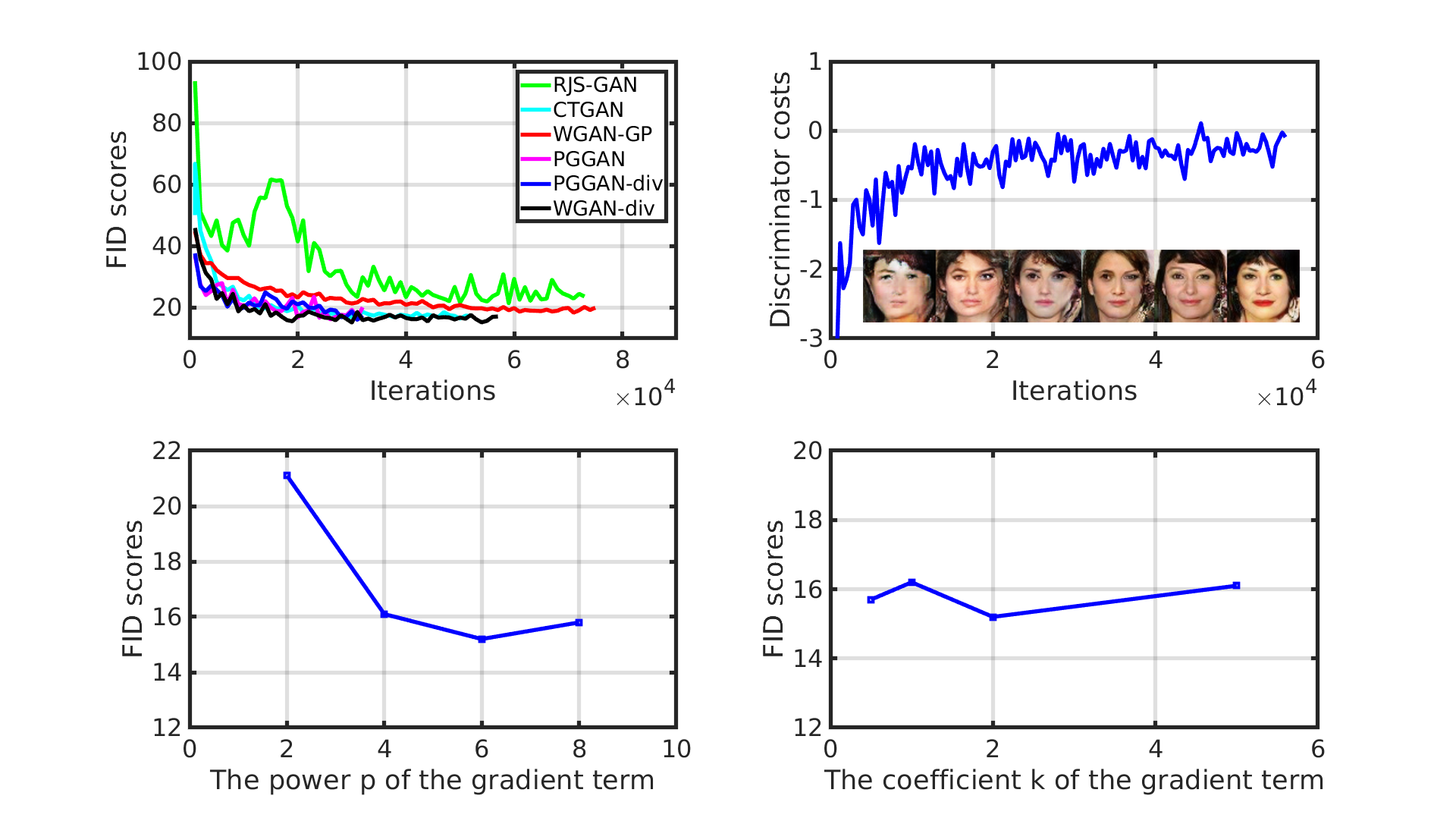}
\end{tabular}
\vspace{-0.2cm}
\caption{Curves of FID vs.\ iteration (top left), Discriminator cost vs.\ iteration (top right), FID vs.\ power $p$ (bottom left), and FID vs.\ coefficient $k$ (bottom right) for WGAN-div on CelebA.}
\label{fig:curve}
\vspace{-0.2cm}
\end{figure}

\section{Experiments}
In this section, we evaluate WGAN-div on toy datasets and three widely used image datasets---CIFAR-10, CelebA~\cite{liu2015deep} and LSUN \cite{yu2015lsun}.
As a preliminary evaluation, we use low-dimensional datasets such as Swiss roll, 8 Gaussians and 25 Gaussians to justify that our proposed W-div can be more effectively  learned  than  W-met used by WGAN-GP and CTGAN,  in  terms  of  more meaningful  value  surfaces  of discriminator $D$ i.e. $f$, and better generated data distribution (Tab.~\ref{tab:3toy}).
Meanwhile, the three large scale datasets highlight a variety of challenges that WGAN-div should address and evaluation on them is adequate to support the advantages of WGAN-div.

Recently,~\cite{heusel2017gans} pointed out that the inception score (IS)~\cite{salimans2016improved} is not reliable because it does not incorporate the statistics of real image samples.
As an alternative, they introduced the Fr\'echet inception distance (FID) to measure the difference between real and fake data distributions. Experiments verified that the FID score is consistent with visual judgment by humans.
Later, \cite{lucic2017gans} conducted a comprehensive study of the state-of-the-art GANs based on FID, 
which confirmed that FID provides fairer assessment.
Hence, we consider the FID score as the major criterion for evaluating our method. 
Also, visual results are provided as a complementary form of verification.

We compare our WGAN-div to the state-of-the-art DCGAN~\cite{radford2015unsupervised}, WGAN-GP~\cite{gulrajani2017improved}, RJS-GAN~\cite{Rothe-IJCV-2016}, CTGAN~\cite{wei2018improve}, SNGAN~\cite{miyato2018spectral}, and PGGAN~\cite{karras2017progressive}.
For each method, we apply the default architectures and hyperparamters recommended by their papers.
The default architectures for $G$ and $D$ of WGAN-div follow the ResNet design~\cite{he2016deep} as presented in Tab.~\ref{tab:net}. 
We use Adam optimization~\cite{kingma2014adam} for updating $G$ and $D$ with a learning rate of $0.0002$ for all three datasets. The number of training steps are 100000 for CelabA and CIFAR-10, and 200000 for LSUN. By cross validation we determine the number of iterations for $D$ per training step to be 4 for CelebA and LSUN, and
5 for CIFAR-10.

\subsection{Hyperparameter Study}
We demonstrate the impact of two important hyperparameters---the power $p$ and the coefficient $k$---on our WGAN-div method.
Both of them control the gradient term of $L_\text{DIV}$.
We report the obtained FID scores on the $64\times64$ CelebA dataset in the bottom row of Fig.~\ref{fig:curve}.
For a fixed optimal $p = 6$ and varying $k$, Fig.~\ref{fig:curve} shows that $L_\text{DIV}$ is not sensitive to changes of $k$, with the FID score fluctuating mildly around 16. 
On the other hand, for a fixed $k = 2$ and changing $p$, we obtain the optimal FID at $p=6$, which differs from the common choice $p=2$ applied in WGAN methods.
The fact that $f_p$ (Eq.~\ref{eq:exist}) converges to the optimal discriminator when $p$ becomes larger may explain why $L_\text{DIV}$ favors a larger power $p$.
To summarize, our default $p, k$ are determined to be $p=6$ and $k=2$.

\subsection{Stability Study}
In this section we evaluate the stability of our method to changes in architecture and compare it to other approaches.  
In this light, we apply various architecture settings for WGAN-div, WGAN-GP, and RJS-GAN, which represent three types of statistical distances:  W-div, W-met, and $\mathfrak{f}$-divergence. 
We train these methods with two standard architectures---ConvNet as used by DCGAN~\cite{radford2015unsupervised} and ResNet~\cite{he2016deep}, which is used by WGAN-GP~\cite{gulrajani2017improved}.
Since batch normalization~\cite{ioffe2015batch} (BN) is considered to be a key ingredient in stabilizing the training process~\cite{radford2015unsupervised}, we also evaluate the FID without BN.
In total, we use four settings: ResNet, ResNet without BN,  ConvNet, and ConvNet without BN.
As shown in Tab.~\ref{tab:stable}, each column reports the visual and FID results obtained under the same architecture. Our WGAN-div achieves the best FID scores for all four settings.
Tab.~\ref{tab:stable} also features corresponding visual results. Compared to WGAN-GP and RJS-GAN, WGAN-div produces more visually pleasing images and the visual quality remains more stable under changing settings.
This experimental study confirms the advantages gained by our W-div and its identical objective $L_\text{DIV}$.

\begin{table}[t]
\centering
\resizebox{1\textwidth}{!}{
\begin{tabular}{ccccc}
\toprule
& ResNet & ResNet without BN & ConvNet& ConvNet without BN \\ \hline
WGAN-GP  & 18.4                                                                           & 20.3                                                                           & 21.2                      & 24.6                                                                           \\ 
& 
\includegraphics[width=0.2\linewidth]{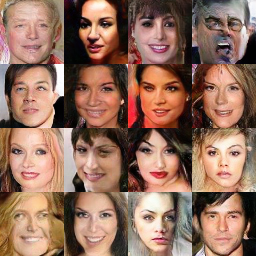}&
\includegraphics[width=0.2\linewidth]{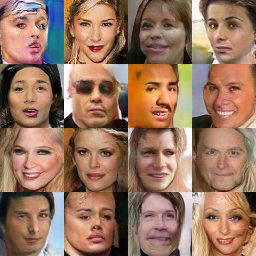}&
\includegraphics[width=0.2\linewidth]{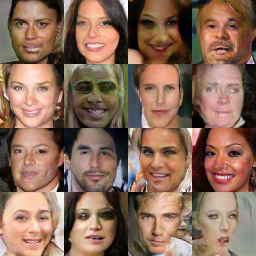}&
\includegraphics[width=0.2\linewidth]{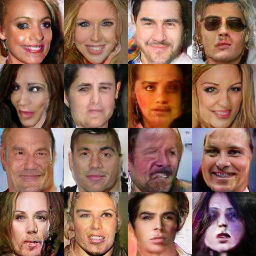}\\ \hline

RJS-GAN & 21.4                                                                 & 23.2   & 21.7  &22.4\\ 

& \includegraphics[width=0.2\linewidth]{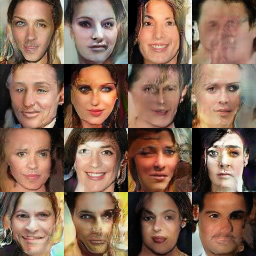}&
\includegraphics[width=0.2\linewidth]{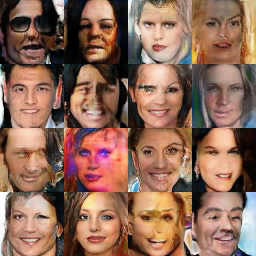}&
\includegraphics[width=0.2\linewidth]{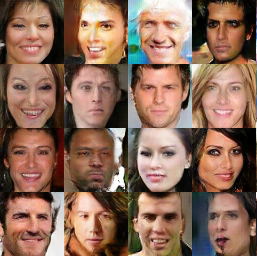}&
\includegraphics[width=0.2\linewidth]{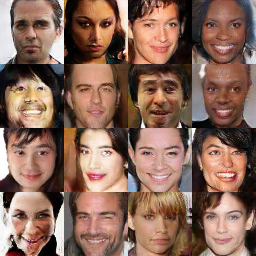}\\ \hline

\textbf{WGAN-div} & \textbf{15.2}                                                                 & \textbf{18.6}                                                                  & \textbf{17.5}            & \textbf{21.5}                                                                  \\ 

& \includegraphics[width=0.2\linewidth]{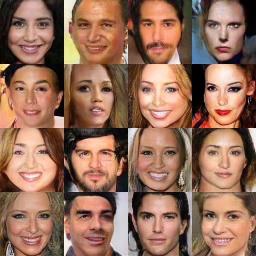}&
\includegraphics[width=0.2\linewidth]{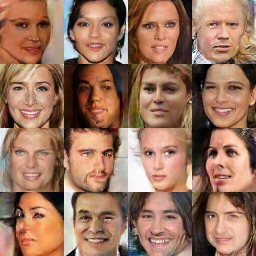}&
\includegraphics[width=0.2\linewidth]{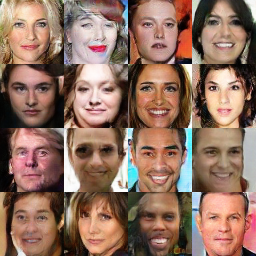}&
\includegraphics[width=0.2\linewidth]{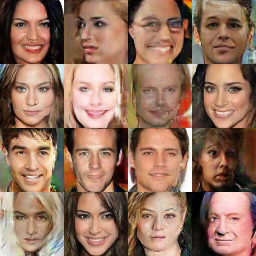}\\ \bottomrule
\end{tabular}
}
\vspace{1mm}
\caption{FID scores and qualitative comparison of various architectures on CelebA.}
\label{tab:stable}
\vspace{-0.4cm}
\end{table}

\begin{figure*}[tb!]
\centering
\begin{tabular}{cccc}
\toprule
CT-GAN & RJS-GAN &WGAN-GP & \textbf{WGAN-div}\\ \hline
\includegraphics[width=0.2\linewidth]{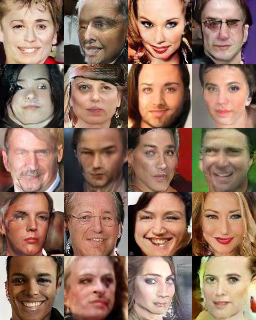}&
\includegraphics[width=0.2\linewidth]{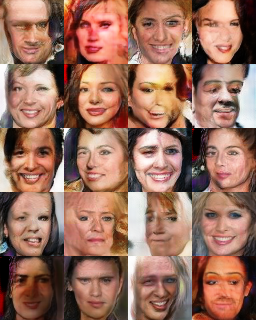}&
\includegraphics[width=0.2\linewidth]{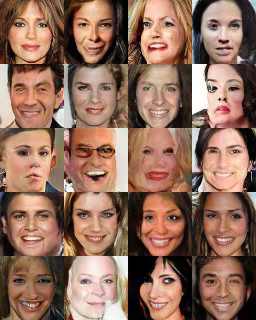}&
\includegraphics[width=0.2\linewidth]{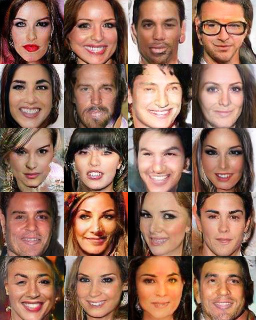}\\

\includegraphics[width=0.2\linewidth]{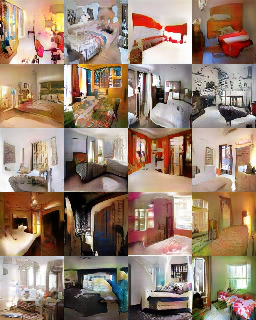}&
\includegraphics[width=0.2\linewidth]{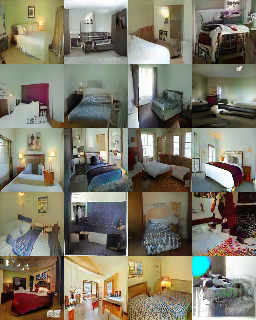}&
\includegraphics[width=0.2\linewidth]{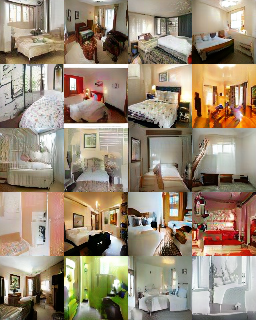}&
\includegraphics[width=0.2\linewidth]{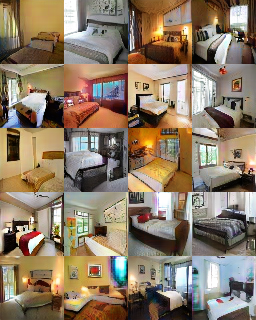}\\

\includegraphics[width=0.2\linewidth]{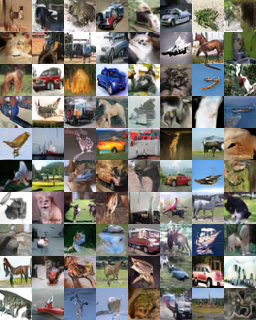}&
\includegraphics[width=0.2\linewidth]{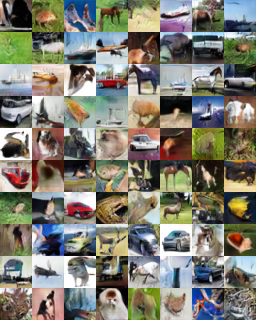}&
\includegraphics[width=0.2\linewidth]{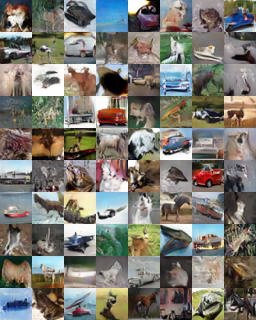}&
\includegraphics[width=0.2\linewidth]{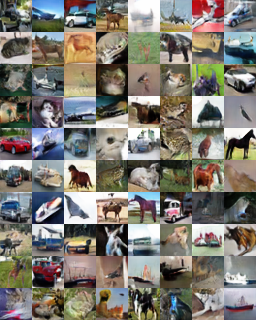}\\
\end{tabular}
\caption{Visual results of WGAN-div and compared methods on CelebA (top row), LSUN (middle row), and CIFAR-10 (bottom row).}
\label{fig:visual}
\vspace{-0.2cm}
\end{figure*}

\subsection{Evaluation on the Standard Training Scheme}
In this experiment, we intend to fairly compare the performance of various GANs by ruling out the impact caused by fine-tuned training strategies.
For this purpose, we follow the standard, i.e.\ non-growing, training scheme, which fixes the size and architecture of the discriminator and generator through the whole training process. 
We compute the FID scores for DCGAN, WGAN-GP, RJS-GAN, CTGAN, and WGAN-div.
The configurations of the compared methods are set according to the recommendations from the authors.
The results are reported in Tab.~\ref{tab:wgan}.
WGAN-div reaches the best FID scores among the compared approaches, which quantitatively confirms the advantages of our method.

While the FID score of WGAN-div mildly outperforms the state-of-the-art methods on the dataset CIFAR-10,
it demonstrates clearer improvements on the larger scale datasets CelebA and LSUN.
Similarly, the facial results shown in Fig.~\ref{fig:visual} tell us that WGAN-div is better than the compared methods with regard to diversity and semantics.
For example, Fig.~\ref{fig:visual} shows diverse faces generated by WGAN-div in terms of gender, age, facial expression and makeup.
We can make the same conclusions on LSUN. 
The proposed WGAN-div outperforms the compared methods with a considerable margin both quantitatively and qualitatively.
For example, WGAN-div achieves an FID score of 15.9 on LSUN, which is 4.4 lower than CTGAN, which is already an improved version of WGAN-GP, that introduced an extra regularizer to enhance WGAN-GP.

The examples of visually plausible bedrooms shown in Fig.~\ref{fig:visual} further highlight the advantages gained by introducing W-div in the GAN model. 
For the interpolation results in the latent space please check our supplementary material.

The top row of Fig.~\ref{fig:curve} reports the learning curve of the compared methods showing that the training process of our WGAN-div is comparatively stable and converges fast. It achieves top FID scores with less than 60K iterations. The top right plot of Fig.~\ref{fig:curve} illustrates the meaningful correlation between image quality and discriminator cost. It is worth mentioning that \cite{heusel2017gans} proposed a two time-scale update method to generally improve the training of a variety of GANs. We believe that WGAN-div can also benefit from such a sophisticated update rule. However, due to the space limit, this is left for further studies.

\begin{table}[!tp]
\centering
\begin{small}
\begin{tabular}{cccc}
\toprule
               & CIFAR-10        & CelebA         & LSUN           \\ \midrule
DCGAN~\cite{radford2015unsupervised}          & 30.9           & 52.0            & 61.1           \\
WGAN-GP~\cite{gulrajani2017improved}        & 18.8          & 18.4          & 26.8            \\
RJS-GAN~\cite{roth2017stabilizing}         & 19.6          & 21.4          & 16.7            \\
CTGAN~\cite{wei2018improve}            & 18.6          & 16.4          & 20.3            \\
SNGAN~\cite{miyato2018spectral}           & 21.7*         & -          & -            \\
\textbf{WGAN-div} & \textbf{18.1} & \textbf{15.2} & \textbf{15.9} \\ \bottomrule
\end{tabular}
\end{small}
\vspace{1mm}
\caption{FID comparison between WGAN-div and the state-of-the-art methods. The result with a * was taken from the original paper~\cite{miyato2018spectral}.}
\label{tab:wgan}
\vspace{-0.4cm}
\end{table}

\begin{table}[!tp]
\centering
\begin{small}
\begin{tabular}{cccc}
\toprule
        & Resolution       & CelebA         & LSUN           \\ \midrule

PGGAN            & $64\times64$         & 16.3          & 17.8            \\ 
\textbf{PGGAN-div}  & $64\times64$  & \textbf{16.0} & \textbf{16.5} \\ \midrule

PGGAN           & $128\times128$           & 14.1          & \textbf{15.4}            \\ 
\textbf{PGGAN-div} & $128\times128$     & \textbf{13.5} & 15.5 \\ \midrule

PGGAN          & $256\times256$              & -          & 15.1            \\ 
\textbf{PGGAN-div} & $256\times256$     & - & \textbf{14.9} \\
\bottomrule
\end{tabular}
\end{small}
\vspace{1mm}
\caption{FID comparison between PGGAN-div and PGGAN at different resolutions.}
\label{tab:pggan}
\vspace{-0.6cm}
\end{table}

\subsection{Evaluation on the Progressive Growing Training Scheme}
Inspired by the success of PGGAN~\cite{karras2017progressive}, which trained a W-met based GAN model in a progressive growing fashion,
we evaluate how our objective $L_\text{DIV}$ performs with this sophisticated training scheme.
More specifically, we replace $L_\text{PG}$ with our $L_\text{DIV}$ while following the default configurations suggested in~\cite{karras2017progressive} and propose PGGAN-div.
However, computing the FID scores for this experimental setting is challenging, as it is non-trivial to adapt existing FID models for evaluating higher resolution generated images.
Since~\cite{karras2017progressive} does not specify the details of how their FID scores were computed for higher resolution images,  
we propose to downscale higher resolution images to $64\times64$ resolution and then compute the FID score.
The resulting scores are reported in Tab.~\ref{tab:pggan}.

Interestingly, Tab.~\ref{tab:pggan} shows that, for low resolution images, the FID score of PGGAN is slightly worse than the one of some top methods reported in Tab.~\ref{tab:wgan}, including WGAN-div.
We believe that this phenomenon is not surprising.
Since it is comparatively easy to learn a data distribution in low dimensional space, applying the standard training scheme suffices to achieve good FID scores.
There is no need to introduce the sophisticated progressive growing strategy during the low dimensional phase.
For higher resolution images ($128\times128$ and $256\times256$) on the other hand, the FID scores for both PGGAN and PGGAN-div decrease with non-negligible margin.
It is worth mentioning that our PGGAN-div slightly improves the FID scores over the original PGGAN,
demonstrating the stability of our objective $L_\text{DIV}$ under a sophisticated training scheme.

\begin{figure*}[p!]
\centering
\begin{tabular}{c}
\includegraphics[width=1.\linewidth]{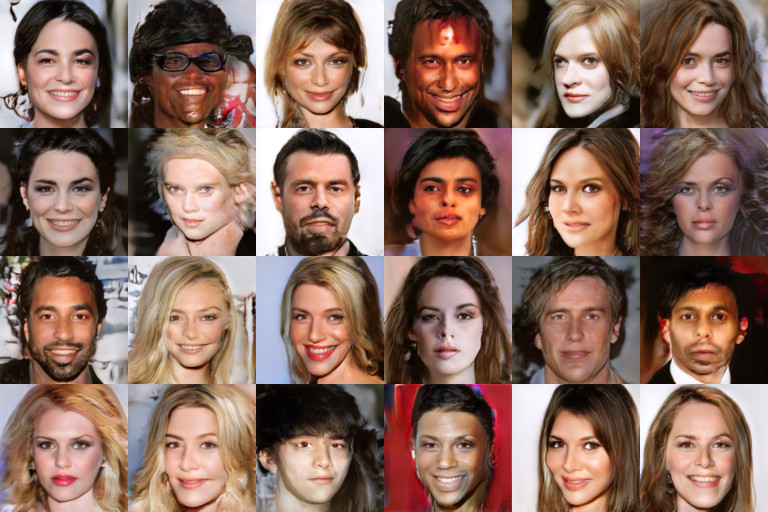}\\
\includegraphics[width=1.\linewidth]{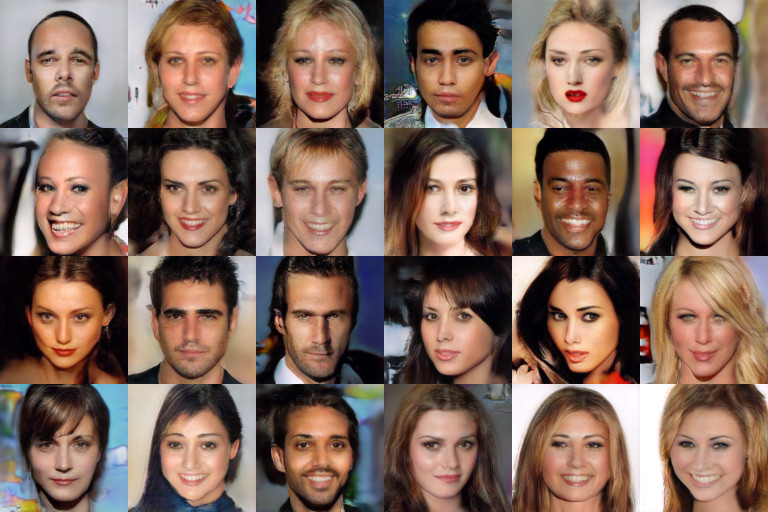}\\
\end{tabular}
\caption{Visual results of PGGAN (top), PGGAN-div (bottom) on CelebA-HQ.}
\label{fig:visual_celeba_HQ}
\end{figure*}

\begin{figure*}[p!]
\centering
\begin{tabular}{c}
\includegraphics[width=1.\linewidth]{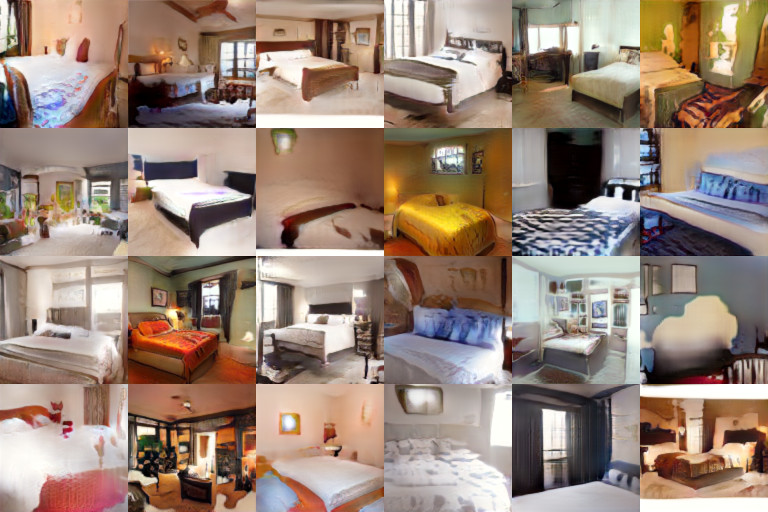}\\
\includegraphics[width=1.\linewidth]{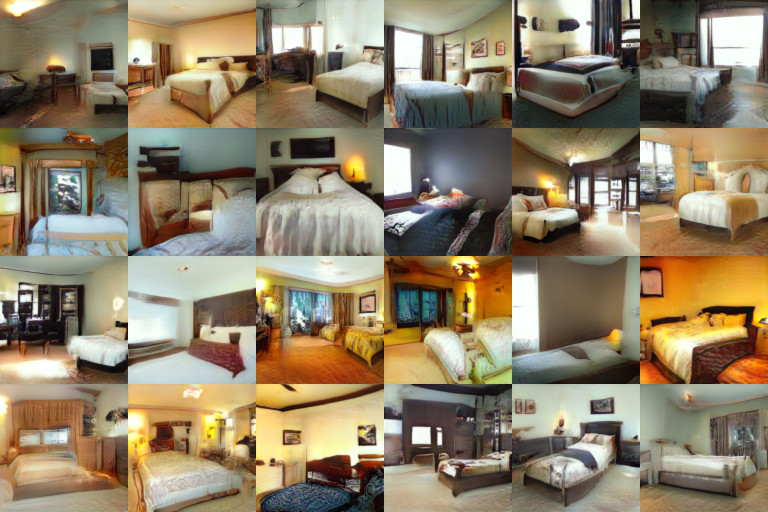}\\
\end{tabular}
\caption{Visual results of PGGAN (top), PGGAN-div (bottom) on $256\times256$ LSUN.}
\label{fig:visual_lsun_256}
\end{figure*}

We also present the $256 \times 256$ visual results for CelebA-HQ (Fig.~\ref{fig:visual_celeba_HQ}) and LSUN (Fig.~\ref{fig:visual_lsun_256}). Since CelebA-HQ was generated by post-processing CelebA~\cite{karras2017progressive}, we do not report its FID scores due to the distribution shift introduced by the artificial post-processing algorithms.
The visual results in Fig.~\ref{fig:visual_celeba_HQ} and Fig.~\ref{fig:visual_lsun_256} demonstrate that our PGGAN-div is very competitive compared to the original PGGAN for both datasets.
To summarize, we demonstrate the stability of our W-div objective under this training scheme. 

\section{Conclusion}
In this paper, we introduced a novel Wasserstein divergence which does not require the $1$-Lipschitz constraint.
As a concrete example, we equip the GAN model with our Wasserstein divergence objective, resulting in WGAN-div.
Both FID score and qualitative performance evaluation demonstrate the stability and superiority of the proposed WGAN-div over the state-of-the-art methods.

\vskip 0.2cm
\noindent\textbf{Acknowledgment.} We would like to thank Nvidia for donating the GPUs used in this work.

\section*{A \quad Proof of Theorem 1}
\paragraph{\textit{Proof.}}
First of all, it is trivial to see that $\mathcal{W}_{p,k}^{'}$ is nonpositive since we can always let $f = 0$. Next, we show identity:

\noindent $``\Rightarrow ":$ Assume two probability measures $\mathbb{P}_r= \mathbb{P}_g$ almost everywhere, then the first two terms of Eq.(\ref{eq:wd}) vanish, that is,
\begin{equation}
\label{eq:id1}
 \underset{\bm{x} \sim \mathbb{P}_r}{\mathbb{E}}[f(\bm{x})] - \underset{\tilde{\bm{x}} \sim \mathbb{P}_{g}}{\mathbb{E}}[f(\tilde{\bm{x}})] = 0.
\end{equation}
Since 
\begin{equation}
\label{eq:id2}
k \underset{\Hat{\bm{x}} \sim \mathbb{P}_{u}}{\mathbb{E}}[\|\nabla f(\Hat{\bm{x}})\|^p] \geq 0,
\end{equation}
and the equality holds if $f = 0$, hence $\mathcal{W}_{p,k}^{'}(\mathbb{P}_r, \mathbb{P}_g) = 0$.

\noindent $``\Leftarrow ":$ Assume $\mathbb{P}_r \neq \mathbb{P}_g$, there exists a  $f \in C_c^{1}(\Omega)$ such that
\begin{equation}
\label{eq:id3}
 \underset{\bm{x} \sim \mathbb{P}_r}{\mathbb{E}}[f(\bm{x})] - \underset{\tilde{\bm{x}} \sim \mathbb{P}_{g}}{\mathbb{E}}[f(\tilde{\bm{x}})] = -\lambda_1 < 0,
\end{equation}
and w.l.o.g
\begin{equation}
\label{eq:id4}
k \underset{\Hat{\bm{x}} \sim \mathbb{P}_{u}}{\mathbb{E}}[\|\nabla f(\Hat{\bm{x}})\|^p] = \lambda_2 > 0.
\end{equation}
We can always find a positive $\lambda$ with $ \lambda^{p - 1} < \frac{\lambda_1}{\lambda_2}$ such that for $f' =  \lambda f$ it satisfies
\begin{equation}
\label{eq:id6}
\begin{aligned}
 \underset{\bm{x} \sim \mathbb{P}_r}{\mathbb{E}}[f'(\bm{x})] - \underset{\tilde{\bm{x}} \sim \mathbb{P}_{g}}{\mathbb{E}}[f'(\tilde{\bm{x}})]
+ k \underset{\Hat{\bm{x}} \sim \mathbb{P}_{u}}{\mathbb{E}}[\|\nabla f'(\Hat{\bm{x}})\|^p] &= \\
 \lambda (\underset{\bm{x} \sim \mathbb{P}_r}{\mathbb{E}}[f(\bm{x})] - \underset{\tilde{\bm{x}} \sim \mathbb{P}_{g}}{\mathbb{E}}[f(\tilde{\bm{x}})])
+ \lambda^{p} k \underset{\Hat{\bm{x}} \sim \mathbb{P}_{u}}{\mathbb{E}}[\|\nabla f(\Hat{\bm{x}})\|^p]  &= \\ 
\lambda(-\lambda_1 + \lambda^{p - 1} \lambda_2) < 0,
\end{aligned}
\end{equation}
taking the infimum over all $f$ we have $\mathcal{W}_{p,k}^{'}(\mathbb{P}_r, \mathbb{P}_g) < 0$.
As for the symmetry, we observe that 
\begin{equation}
\begin{aligned}
\mathcal{W}_{p,k}^{'}(\mathbb{P}_r, \mathbb{P}_g) &= \underset{f \in C_c^{1}(\Omega)}{\mathrm{inf}}  \underset{\bm{x} \sim \mathbb{P}_r}{\mathbb{E}}[-f(\bm{x})] - \underset{\tilde{\bm{x}} \sim \mathbb{P}_{g}}{\mathbb{E}}[-f(\tilde{\bm{x}})] 
+ k \underset{\Hat{\bm{x}} \sim \mathbb{P}_{u}}{\mathbb{E}}[\|\nabla -f(\Hat{\bm{x}})\|^p]\\
&= \underset{f \in C_c^{1}(\Omega)}{\mathrm{inf}}  \underset{\tilde{\bm{x}} \sim \mathbb{P}_g}{\mathbb{E}}[f(\tilde{\bm{x}})] - \underset{\bm{x} \sim \mathbb{P}_{r}}{\mathbb{E}}[f(\bm{x})]
+ k \underset{\Hat{\bm{x}} \sim \mathbb{P}_{u}}{\mathbb{E}}[\|\nabla f(\Hat{\bm{x}})\|^p] \\ 
&= \mathcal{W}_{p,k}^{'}(\mathbb{P}_g, \mathbb{P}_r).  \qed
\end{aligned}
\end{equation}

\newpage

\section*{B \quad Study of Sampling Strategies}
\begin{table*}
\centering
\resizebox{1\textwidth}{!}{
\begin{tabular}{ccccccc}
\toprule
& (1) & (2)  & (3) & (4) & (5) & (6)\\ \hline
WGAN-GP  & 18.4 & 20.1  & 19.3  & 19.0  & 18.3  & 17.0         \\ 
& 
\includegraphics[width=0.2\linewidth]{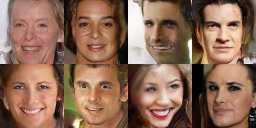}&
\includegraphics[width=0.2\linewidth]{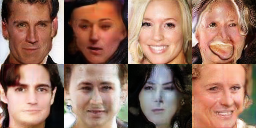}&
\includegraphics[width=0.2\linewidth]{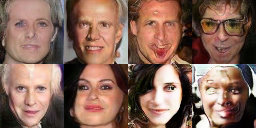}&
\includegraphics[width=0.2\linewidth]{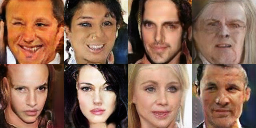}&
\includegraphics[width=0.2\linewidth]{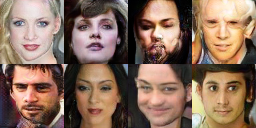}&
\includegraphics[width=0.2\linewidth]{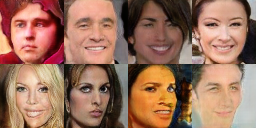}\\ \hline

CTGAN & 16.4  & 16.5  &17.7 & 17.2 & 17.9 & 17.5\\ 

& \includegraphics[width=0.2\linewidth]{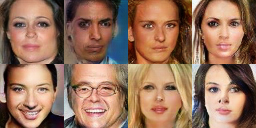}&
\includegraphics[width=0.2\linewidth]{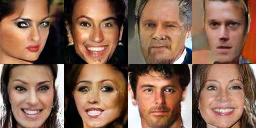}&
\includegraphics[width=0.2\linewidth]{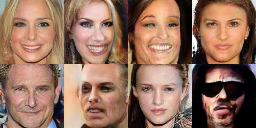}&
\includegraphics[width=0.2\linewidth]{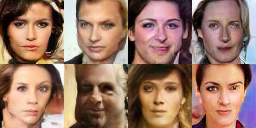}&
\includegraphics[width=0.2\linewidth]{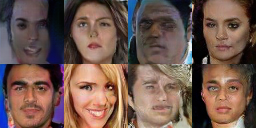}&
\includegraphics[width=0.2\linewidth]{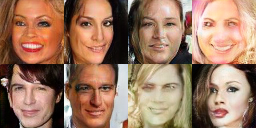}\\ \hline

\textbf{WGAN-div} & \textbf{15.2} & \textbf{15.9} & \textbf{15.1} & \textbf{14.5}  & \textbf{15.5} & \textbf{14.9}\\ 

& \includegraphics[width=0.2\linewidth]{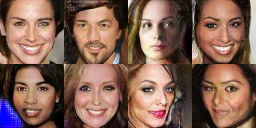}&
\includegraphics[width=0.2\linewidth]{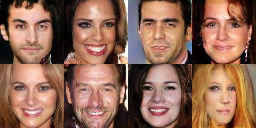}&
\includegraphics[width=0.2\linewidth]{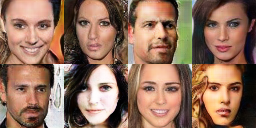}&
\includegraphics[width=0.2\linewidth]{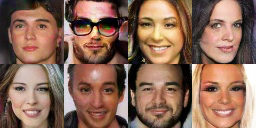}&
\includegraphics[width=0.2\linewidth]{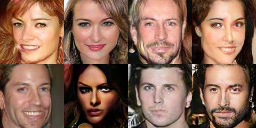}&
\includegraphics[width=0.2\linewidth]{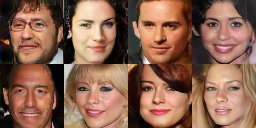}\\ \bottomrule
\end{tabular}
}
\label{tab:samp}
\caption{Visual and FID comparison of various sampling strategies on CelebA.}
\vspace{-0.7cm}
\end{table*}

We study the effects of different sampling strategies corresponding to six types of $\mathbb{P}_u$, that is, 
sampling the linear combination of a real and fake data point~(1),  sampling the linear combination of two real or two fake data points~(2), sampling the linear combination of two data points from both real or fake data~(3), sampling both real or fake data points~(4), sampling fake data points~(5) and sampling real data points~(6).
Tab.˜\ref{tab:samp} shows that our WGAN-div consistently outperforms compared methods regardless of the sampling strategy. These results also confirm that the selection of $\mathbb{P}_u$ plays a minor role for the final performance. In the paper, we thus empirically determine $\mathbb{P}_u$ using sampling strategy~(1).

\newpage

\section*{C \quad Interpolation Results}

\begin{figure*}[]
\centering
\begin{tabular}{cc}
\includegraphics[width=0.5\linewidth]{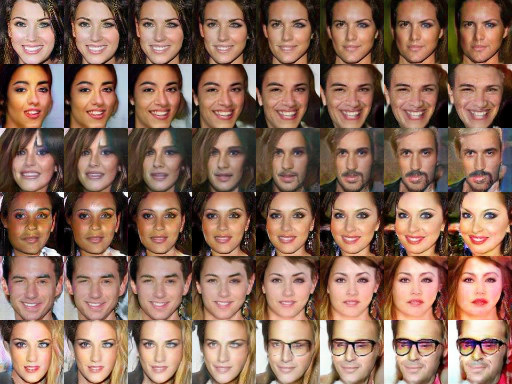}&
\includegraphics[width=0.5\linewidth]{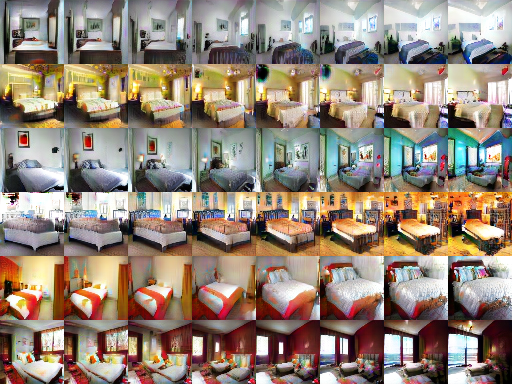}\\
\end{tabular}
\caption{Interpolation results of $64 \times 64$ CelebA and LSUN.}
\vspace{-0.7cm}
\end{figure*}

\bibliographystyle{splncs}

\end{document}